\pdfoutput=1
\documentclass[11pt]{article}

\usepackage[]{EACL2023}

\usepackage{times}
\usepackage{latexsym}

\usepackage[T1]{fontenc}
\usepackage[utf8]{inputenc}

\usepackage{microtype}

\usepackage{inconsolata}
\usepackage{microtype}
\usepackage{graphics}
\usepackage{graphicx}
\usepackage{booktabs}
\usepackage{multirow}
\usepackage{tablefootnote}
\usepackage{float}
\usepackage{placeins}
\usepackage{amsmath}

\newcommand\mwide{M25$_{wide}$}
\newcommand\mdeep{M25$_{deep}$}
\newcommand\mtwide{Mt25$_{wide}$}
\newcommand\mtdeep{Mt25$_{deep}$}

\title{Efficiently Upgrading Multilingual Machine Translation Models \linebreak to Support More Languages}

\author{
  Simeng Sun$^1$\thanks{\ \ Work done during an internship at Meta AI.}\hspace{3mm}Maha Elbayad$^2$\hspace{3mm} Anna Sun$^2$\hspace{3mm} James Cross$^2$\thanks{\ \ Work done while working at Meta AI.} \\
  University of Massachusetts Amherst$^1$ \hspace{1em} Meta AI$^2$ \hspace{1em} \\
  \texttt{simengsun@umass.edu} \\
  \texttt{\{elbayadm, annaysun, jcross\}@meta.com} \\
  }

\begin{document}
\maketitle

\begin{abstract}

With multilingual machine translation (MMT) models continuing to grow in size and number of supported languages, it is natural to reuse and upgrade existing models to save computation as data becomes available in more languages.
However, adding new languages requires updating the vocabulary, 
which complicates the reuse of embeddings. The question of how to reuse existing models while also making architectural changes to provide capacity for both old and new languages has also not been closely studied.
In this work, we introduce three techniques that help speed up effective learning of the new languages and alleviate catastrophic forgetting despite vocabulary and architecture mismatches.
Our results show that by (1) carefully initializing the network, (2) applying learning rate scaling, and (3) performing data up-sampling,
it is possible to exceed the performance of a same-sized baseline model with 30\% computation and recover the performance of a larger model trained from scratch with over 50\% reduction in computation.
Furthermore, our analysis reveals that the introduced techniques help learn the new directions more effectively and alleviate catastrophic forgetting at the same time. We hope our work will guide research into more efficient approaches to growing languages for these MMT models and ultimately maximize the reuse of existing models.

\end{abstract}
\section{Introduction}

% intro outline

% state of the art of multilingual machine translation
Research into multilingual machine translation (MMT)~\cite{aharoni-etal-2019-massively,fan2020englishcentric} has shifted from a relatively small number of translation directions~\citep{dong2015multi,firat2016multi,ha2016toward} to much larger scale, recently reaching up to tens of thousands of translation directions~\citep{nllb-paper,google-next-1k-lang}. Despite the notable increase in the number of supported languages, these models still need to be upgraded as increasing amount of data in new languages are becoming available. 
% the obstacles of adding new languages, the necessity of speeding up adding new languages
The process of adding new languages to existing models is an instance of continual learning~\citep{ring1994continual}, in which the models need to effectively learn new tasks (new translation directions) while not catastrophically forgetting~\citep{NIPS1993_28267ab8,mccloskey1989catastrophic} the knowledge about tasks from the previous training stage (original translation directions).
% Since MMT models are multi-task learners where each language direction represents a task, the process of updating an existing MMT model is essentially an instantiation of continual learning~\citep{} (also called lifelong learning~\citep{} and sequential learning~\citep{}) under a wider context of multi-task learning. Unlike simply fine-tuning a pre-trained models on target tasks~\citep{}, this line of work pays extra attention to countering , that is retaining as much knowledge learned from the previous training phase as possible.

Unlike the most conventional continual learning setup where the model remains the same as the initial learning stage, growing languages for MMT models needs to deal with \emph{new} parameters.
% \maha{Is this specific to MMT or NLP in general? should we also include related work in language modeling?} 
% \jccomment{This should emphasize adding new languages, and therefore needing to support an expanded vocabulary}
Adding new languages to existing training data shifts subword token distribution (e.g., tokens from unseen scripts are added), hence the need to re-train the tokenizer, which adds new embedding parameters to the MMT model.
Previous studies have shown the effectiveness of adapting embedding parameters to retain performance on old translation directions~\citep{Lakew2018TransferLI,escolano-etal-2019-bilingual,garcia-etal-2021-towards}. This is usually done by reusing the embeddings for tokens that overlap between the old and new vocabularies.

One aspect that has not been extensively addressed in previous research is how to deal with other architecture mismatches that may arise during the continual learning phase. When growing MMT models to support many additional languages, it also makes sense to increase the model size overall. This extra capacity can be used to learn not only the new directions well but also the old directions better. It is not obvious, however, how to reuse the parameters from existing models (i.e., to engage in continual learning) given such architectural changes. Thus, in addition to dealing with different vocabularies, we also investigate how to make use of previously trained models when scaling up model size in order to train much more efficiently than training from scratch. 

% However, one aspect missing in previous research is the potential architectural mismatches that arise during the continual learning phase. Old models are trained on fewer number of directions and could have smaller architectures, hence less capacity. Thus it is natural to increase the model sizes in the new learning stage so that the model learns not only the new directions well but also the old directions better. However, in presence of significant architectural changes, the most common way is train the model from scratch on the combination of old and new data, which is resource-consuming, especially for large-scale multilingual models. Therefore, in addition to dealing with different vocabularies, we also investigate how to efficiently reuse the old models when significant architectural mismatches exist between the initial and continual learning stages.

\begin{figure*}
    \centering
    \includegraphics[width=0.99\textwidth]{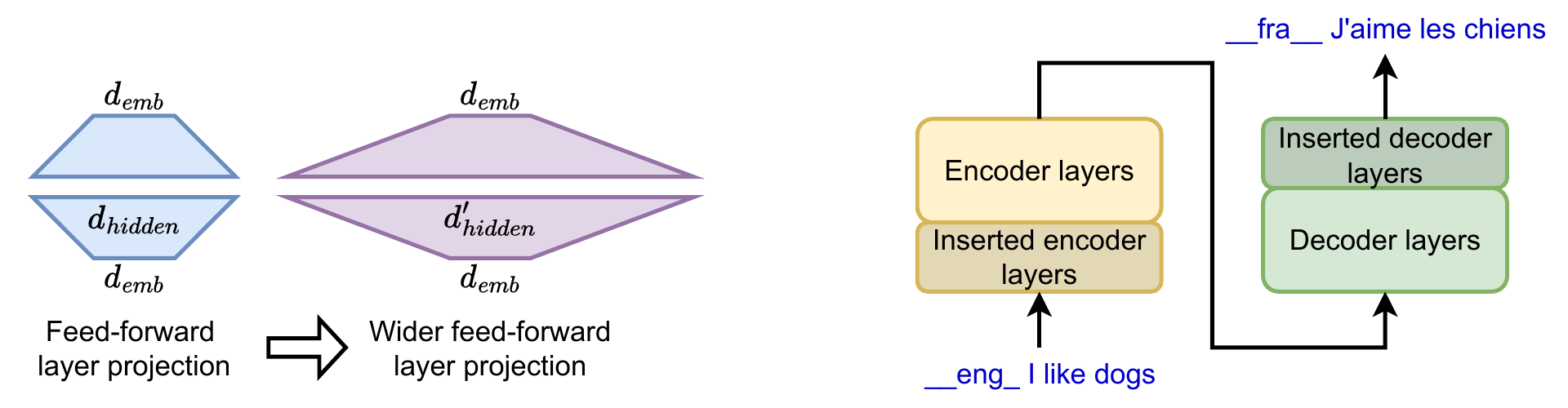}
    \caption{Illustration of two architectural mismatches we tackle in this paper. \textbf{Left:} The hidden dimension in the feed-forward layers is doubled ($d'_{\text{hidden}}=2d_{\text{hidden}}$) during the continual learning stage so that the model becomes ``wider''. \textbf{Right:} Additional layers are inserted at the bottom of encoder and the top of decoder so that the model becomes ``deeper''. Both architectural changes increase capacity and are not well-addressed by previous works.}
    \label{fig:archi_mismatch_illus}
\end{figure*}

In this work, we introduce a recipe that helps significantly reduce the amount of required computation for continually learning new languages in MMT models. The recipe involves training the models with a combination of three techniques: (1) careful initialization of the network (2) applying learning rate scaling and (3) up-sampling selected language pairs. 
% \maha{not only new pairs}
We validate our method on settings both with and without \emph{architecture mismatches} (e.g., models becoming wider or deeper as shown in Figure~\ref{fig:archi_mismatch_illus} to have extra capacity for both old and new directions). We compare our models with strong same-sized baselines that are trained on all data from scratch.
Our experimental results show that, without architecture mismatches, it is possible to outperform the baseline with just 30\% of computation required by the baseline.
% \maha{Original training time is unclear, is it M20 or M25 training time - I understand that in our setup it's 30K updates regardless}
When training larger models, 
% \maha{remove the e.g. since ffn dim hasn't been introduced yet?}
less than 50\% of the original computation is required to match the full performance of the wider baseline model, and less than 10\% of computation is needed to recover over 95\% of the corresponding baseline performance. We further conduct a suite of analysis which shows that: 
\begin{itemize}
    \item Proper initialization of the parameters before continual learning is crucial for fast convergence.
    \item Data up-sampling is vital to achieving good performance on new language pairs.
    \item Scaling down the learning rate for reused parameters helps alleviate catastrophic forgetting.
\end{itemize}

It is our hope that this work will help save computation for future research into large-scale multilingual machine translation, guide more efficient reuse of existing models for continual learning, and allow people to efficiently adapt large publicly-released MMT models for new languages and datasets.

% \maha{Add a sentence or two about the paper's outline.}
% what we propose
% our contribution

% \sscomment{probably move this part to introduction}

% Emphasize that adding new languages is different from conventional fine-tuning or continual learning because of vocabulary and architectural mismatches. Due to these differences, models usually need to be trained from scratch, which is resource-consuming if the data set and the number of languages are large.

% \sscomment{TODO: make it clear that architectural mismatch also includes vocabulary mismatch}

% \sscomment{TODO: articulate why we don't compare with adapter}
% \paragraph{Vocabulary mismatch} (adding new languages, especially new scripts, requires retraining the SPM tokenizer, and thus new vocabulary)

% \paragraph{Architectural mismatch} (increasing the model capacity is a natural way to support new languages. we investigate two directions: (1) wider ffn (2) deeper network. There are other configurations e.g. increase both hidden dim and num layers, add adapters, MoE models, they are out of scope of this project.)

\begin{table*}[ht]
    \centering
    \scalebox{0.9}{
    \begin{tabular}{@{}llc|cc|cc|cc@{}}
\toprule
&     & \hspace{0.5em} \textbf{M20} \hspace{0.5em} & \hspace{0.5em} \textbf{M25$^{@100k}$}\hspace{0.5em} & \textbf{Mt25$^{@30k}$} & \textbf{M25$_{wide}^{@100k}$} & \textbf{Mt25$_{wide}^{@50k}$} & \textbf{M25$_{deep}^{@100k}$} & \textbf{Mt25$_{deep}^{@50k}$} \\ \midrule
\multicolumn{1}{l}{\textbf{spBLEU}} & All & 33.4  & 31.6  & 31.8    & 32.8 & 32.8  & 32.8 & 32.4          \\
& Orig. & 33.4        & 33.4        & 33.5    & 34.5 & 34.25   & 34.5 & 33.9          \\
& Added & -            & 24.6        & 25.2    & 25.9 & 27.2  & 26.2 & 26.2          \\ \midrule
\multicolumn{1}{l}{\textbf{chrF++}}   & All & 76   & 74   & 74     & 75  & 75   & 75  & 75           \\
& Orig. & 76         & 76         & 76    & 76  & 76   & 76  & 76           \\
& Added & -            & 67         & 69     & 68  & 70   & 68  & 69           \\ \bottomrule
\end{tabular}
}
    \caption{While continually learning new languages by bootstrapping from a model trained on 20 languages (M20), given new embedding parameters (vocabulary mismatch), one can exceed the performance of a baseline model trained on all languages from scratch (M25$^{@100k}$) with just 30\% computation (Mt25$^{@30k}$). With architectural mismatches, employing our method, the wider-model baseline performance (M25$_{wide}$) and over 98\% performance of a deeper-model baseline (M25$_{deep}^{@100k}$) can be recovered with half of the corresponding baseline computation (Mt25$_{wide}^{@50k}$ and Mt25$_{deep}^{@50k}$ respectively). The ``Orig.'' row shows performance on old 20 languages and the ``Added'' on newly added 5 languages.
    % \maha{add what all, orig and added mean}
    % \maha{use 1 digit precision as in the rest of the tables}
    }
    \label{tab:main_res}
\end{table*}

\section{Method} \label{sec:method}

Adding new languages to existing models, especially languages in new scripts, leads to different subword tokenization,
% \maha{maintaining the same vocabulary size is not the issue, we would have different vocabularies regardless of the size} 
thus different vocabularies, which precludes simple fine-tuning with the exact same model on additional data. 
% More often than not, the old models are of lower capacity 
% \jccomment{I tried to briefly address this in the intro, but it's really our decision to scale to a larger model, and one I think we need to explain/defend in a consistent way...}. 
% Besides enabling new models to translate new directions, one may also want to grow the model capacity so that all directions can be improved. 
During the continual learning stage, we may also want to increase the model size overall to have extra capacity to learn the new languages and improve old languages at the same time.
Therefore, we also investigate two typical architectural changes commonly done to increase the network capacity: (1) make the model ``wider'' by expanding the hidden dimension of the feed-forward layers, and (2) make the model ``deeper'' by inserting new layers to both encoder and decoder.
In this section, we delineate three techniques that we found most effective in achieving computation reduction for the continual learning of MMT models. 
% \maha{the term efficient continual learning is ambiguous, is it computationally efficient or a trade-off between quality and compute ... the term was previously mentioned in the intro and abstract but it's worth specifying what we mean by efficient in this context}

% \jccomment{Maybe say something like: while we tried various approaches (see appendix), the following subsections summarize the three techniques we found most effective:}

% \maha{needs a clarification: there might be some tokens in the old vocab that do not exist in the new one (newly trained spm leading to different merges and so on) - it means that we cannot copy **all** the weights of the old model}
\paragraph{Proper initialization.} While we can copy weights from the old model\footnote{The weights that can be directly copied without any modification include (1) the token embeddings that overlap between old and new model, (2) same-shaped non-embedding parameters.} to the new model, it is not immediately clear how the new parameters (e.g., new token embeddings, new feed-forward weights, new layers) should be initialized and co-adapted with the old weights such that maximal knowledge about the old directions is retained.
% \maha{what do you mean by co-adapted?}
% \maha{another nitpick: the idea of maximal knowledge is ill-defined; I can argue that **maximal** knowledge is not **optimal** knowledge for the new set of tasks} 
Instead of initializing the new parameters randomly, we find that initializing the new embeddings with that of \texttt{<unk>} leads to the best performance. When the network becomes ``wider'' in the continual learning phase, concatenating each old weight matrix with a noisy version of itself performs better than other methods we tried. When the network becomes ``deeper'', initializing new layers with averaged weights of old layers results in slight improvement over other naive initialization methods. For each setup, we compare different initialization methods in section~\ref{subsec:init}.
% \maha{This sentence tries to give technical details but it's not clear enough. I would recommend adding some definition of what we call wider and deeper architectures before this paragraph. Maybe an illustration or at least a reference to figure 1.} 

\paragraph{Data up-sampling.} Similar to~\cite{garcia-etal-2021-towards}, we introduce the new tasks by mixing old and new training data together. Since the main goal of the continual learning phase is to quickly learn the new directions, we up-sample the new pairs so the model gets more learning signals from these new directions. To increase the transfer across related language pairs and reduce the chance of catastrophically forgetting less represented directions in the original training data, we also up-sample a small number
% \maha{is it a small number or all old low-res from the same families as the new ones? I would add / highlight these languages in Table 9 of the appendix as they are in Table 2. We should also define in the appendix what we consider low, ex-low... and specify that ex-low is a subset of low so that when we set a rule for low-res languages it also covers ex-low.}
of the old low-resource languages that are from the same language family as the new languages. We present the effect of  up-sampling these selected directions in section~\ref{subsec:upsample}.
% The upsample ratio for these new directions is a hyperparameter that needs to be tuned first \jccomment{We need to be careful here, as tuning hyperparameters that apply to an entire training run undercuts the message that these techniques save compute}, whose effect we present in the analysis in section~\ref{subsec:upsample}.

\paragraph{Learning rate scaling.} To further mitigate the frequent catastrophic forgetting problem exhibited in continual learning~\cite{ring1994continual,thompson-etal-2019-overcoming}, we scale the learning rate for individual parameters depending on whether they are copied from the old model or not. Based on the assumption that the model better retains the knowledge about old tasks if the weights stay close to that of the old model~\citep{EWC},
% \maha{Was this assumption mentioned in some previous work? if not we should rephrase to make it clear that it's our assumption - that I guess we test later in the analysis section}
we scale down the learning rate for these old parameters while maintaining or scaling up the learning rate for the newly added parameters. In contrast to other methods that incur extra computation such as Fisher Information based loss~\citep{thompson-etal-2019-overcoming}, 
% \maha{Ambiguous. Can we explain what additional computation these methods have e.g. computing the Fisher information matrix}
our approach is simple, straightforward and efficient in alleviating catastrophic forgetting. We present a deeper analysis of learning rate scaling in section~\ref{subsec:lr_scale} and section~\ref{subsec:catastrophic}.

\section{Experiments}

\paragraph{Languages.} We conduct our experiments on 25 languages covering ten language families, four resource levels (high, mid, low, and very low) and four scripts\footnote{We include more details about these languages in Appendix~\ref{sec:language_selection} Table~\ref{tab:language_details}.}. To grow the languages atop existing learned languages, we train the seed model on 20 languages (40 English-centric directions) and add 5 new languages (10 English-centric directions) during the continual learning stage. Mimicking the common scenario where newly added languages are mostly low-resource, we select five low and very-low resource languages covering four language families and four scripts as the new languages to add to the seed model. To verify the validity of our approach, we also experiment with two other 20/5 groupings and one 12/13 division of old/new languages in section~\ref{subsec:seed_lang}\footnote{More language grouping details are shown in Table~\ref{tab:lang_breakdown_new} and Table~\ref{tab:m12t25} in Appendix~\ref{sec:language_selection}.}.
Table~\ref{tab:langbreakdown} shows the groups of original and added languages used in most of our experiments. 
For all our models, we train \texttt{sentencepiece}~\citep{kudo-2018-subword}\footnote{\url{https://github.com/google/sentencepiece}} tokenizers with 64K tokens and a sampling temperature of 2, on joined (source and target) data from all language directions available in each setup.
This means that tokenizers in the initial training on the the old languages, were not exposed to unseen languages.
The data we used are subsets of data used by~\citet{nllb-paper} without any mined or back-translated examples.

\begin{table}[!t]
    \centering
    \begin{tabular}{@{}cccc|c@{}}
\toprule
\multicolumn{4}{c|}{\textbf{Original}} & \textbf{Added} \\ \midrule
lav   & rus   & fin   & spa   & xho   \\
lit   & hin   & est   & deu   & guj   \\
swh   & mar   & pol   & zul$^*$   & npi   \\
ukr   & mkd   & ces   & msa$^*$   & ind   \\
bel   & bul   & fra   & kir$^*$   & kaz   \\ \bottomrule
\end{tabular}
    \caption{The M20 model is trained on the ``Original'' languages. Mt25, \mtwide\ and \mtdeep\ bootstrap from M20 and train on the combination of ``Original'' and ``Added'' languages. We perform data up-sampling over added data in conjunction with related old low-resource languages (marked with $^*$).
    }
    \label{tab:langbreakdown}
\end{table}

\paragraph{Models.} Our baseline models have a standard 12-layer encoder and 12-layer decoder Transformer architecture~\citep{vaswani2017attention} with 16 attention heads, embedding dimension 1024 and hidden dimension 4096. We refer to the baseline model trained on 20 languages and 25 languages from scratch as M20 and M25 respectively. For wider models, we double the hidden dimension size to 8192; and for deeper models, we insert 6 extra layers to encoder and decoder each. We refer to the baseline wide and deep models trained on 25 languages from scratch as \mwide\ and \mdeep\ respectively, and refer to the models which bootstrap from the M20 model and trained to support new languages while remaining the same architecture, using a wider network, and deeper network as Mt25, \mtwide\ and \mtdeep\ respectively.  All translation directions were evaluated on FloRes~\cite{flores_1,flores_2} dev set and use \texttt{spBLEU}\footnote{\url{https://github.com/facebookresearch/flores/tree/main/previous_releases/flores101\#spbleu-evaluation}}
% \maha{specify if it's FLORES-200's spBLEU or  FLORES-101 since all directions are part of 101 and both spms are valid}
dand \texttt{chrF++}~\citep{%
%popovic-2015-chrf,
popovic-2017-chrf%
} to evaluate the best checkpoint selected by validation loss with beam size 4.\footnote{ We use these two metrics instead of BLEU or pre-trained metrics ~\citep{kocmi-etal-2021-ship} because over half of the languages we use are of mid- to low-resource and ~\citet{flores_1} show spBLEU treats low-resource languages more fairly.}
% \maha{Does that mean that we are reporting the results of checkpoint\_best which is potentially trained for less than 30k/50k updates??}
% \maha{I don't see why we have to cite chrf if we're only reporting chrf++. chrf++ is usually reported on a 0-100 scale. If you're using scarebleu, can you add the score signatures}
% \maha{add beam search width}
% \sscomment{anything to add?}\jccomment{Is it chrf++ ? We should cite either way}

\paragraph{Training details} We train baseline models (M25, \mwide, \mdeep) for 100K updates with batch size $\sim$444K tokens peak learning rate 0.003 warmed up with 8000 steps. For the bootstrapped models, we train Mt25 for 30k updates and \mtwide\  and \mtdeep\ for 50k updates since there are more added parameters. We use temperature 1 and prepend encoder/decoder language token at the beginning of each example.
% \maha{Add Temperature of sampling, encoder/decoder-langtok and any other MMT specific params. Are we planning to open-source the code?}
% \maha{can you add parameter counts in some of the existing tables or add the percentage of new parameters in each of the three models Mt25, \mtwide and \mtdeep}
All models are trained with %dropout~\citep{10.5555/2627435.2670313} 0.0, 
attention dropout 0.1 and label smoothing with $\epsilon=0.1$~\citep{7780677}.
The baseline models M25, \mwide\ and \mdeep\ complete training of 100K updates on 64 GPUs in $\sim$24h, $\sim$29.9h, and $\sim$29.0h respectively. We do not reset the learning rate scheduler for the second training phase.%
\footnote{We tried resetting learning scheduler (i.e., learning rate warms up 8K steps from 0 to 0.003 and then starts to decrease), small constant learning rate, and not resetting learning rate scheduler in our preliminary experiments and the last leads to the best results.}
The data of selected languages 
are up-sampled by 5 
and the learning rate of the old parameters in \mtwide\ is multiplied by 0.5 and in \mtdeep\ by 0.05 at the beginning while linearly increasing to 0.5. We present the effect of these hyperparameters in the next section.

\begin{table}[!t]
    \centering
   \scalebox{0.99}{
   \begin{tabular}{@{}lccc@{}}
\toprule
          & \multicolumn{1}{l}{\textbf{All}} & \textbf{Orig.}         & \textbf{Added}         \\ \midrule
\textbf{Mt25}   & \textbf{31.8}       & \textbf{33.5}     & \textbf{25.2}     \\
\hspace{2em} random init all       & 28.2$^*$       & 29.4$^*$     & 23.4$^*$     \\
\hspace{2em} random init new       & 31.7       & 33.5     & 24.9$^*$     \\
\hspace{2em} no up-sampling    & 31.3$^*$       & 33.4     & 22.6$^*$     \\
\hspace{2em} no lr scaling       & 31.5$^*$       & 33.1$^*$     & 25.1     \\ \midrule
\textbf{Mt25$_{wide}$} & \textbf{32.6}       & \textbf{34.1}     & \textbf{26.3}     \\
\hspace{2em} random init all       & 26.7$^*$       & 28.0$^*$     & 21.5$^*$     \\
\hspace{2em} random init new       & 29.9$^*$       & 31.3$^*$     & 24.6$^*$     \\
\hspace{2em} no up-sampling    & 31.9$^*$       & 34.1     & 23.3$^*$     \\
\hspace{2em} no lr scaling       & 32.3$^*$       & 33.9$^*$     & 26.2$^*$     \\ \midrule
\textbf{Mt25$_{deep}$} & \textbf{32.2}       & 33.8     & 25.6     \\
\hspace{2em} random init all       & 30.5$^*$   & 31.7$^*$ & \textbf{25.7} \\
\hspace{2em} random init new       & 32.2       & \textbf{33.9}     & 25.6     \\
\hspace{2em} no up-sampling    & 31.2$^*$       & 33.5$^*$     & 22.2$^*$     \\
\hspace{2em} no lr scaling       & 31.6$^*$       & 33.0$^*$     & 26.1     \\ \bottomrule
\end{tabular}
   }
    \caption{Ablation of the effect of not using M20 weights (random init all),
    % \maha{doesn't random init all mean that we don't use the weights of M20 at all? - the expression "ablation of the effect" is confusing. If we want to use consistent notation, then random init all/new should be worded with 'no'}
    the effect of not carefully initializing new parameters (random init new), 
    not having data up-sampling and not applying learning rate scaling across three scenarios. All numbers are spBLEU scores of models evaluated after 30K updates. $^*$ indicates $p$-value of paired T-test against corresponding best model (Mt25, \mtwide, \mtdeep) is smaller than $0.05$.
    % \maha{Mention that these are spBLEU scores + in other captions when relevant}
    % \jccomment{The "$-$" signs in the ablation table look like they mean minus (which is why this is often used in ablation tables to list things you are taking away). Since these do not mean that exactly (the word "no" is used instead) maybe we should consider another format?}
    }
    \label{tab:ablation}
\end{table}

\paragraph{Main results.} Results in Table~\ref{tab:main_res} indicate that by properly applying the techniques introduced earlier, it is possible to recover the baseline performance with significantly less computation. When architecture remains the same during the second learning phase, better overall performance can be achieved (31.8 vs. 31.6) after 30\% of the updates required by the baseline model M25. The gains come from effective learning of both the old and new directions while the latter seems to be better learned than the baseline model (25.2 vs. 24.6).  While training into a wider model, applying our techniques recovers the performance with approximately half of the \mwide\ computation. Although we did not fully recover the performance while training into a deeper model, over 98\% of the baseline performance can be achieved using our techniques with half of the baseline \mdeep\ computation. We find that the major degradation in \mtdeep\ comes from the worse performance on the original directions (33.9 vs. 34.5) which suggests mitigating catastrophic forgetting is harder when expanding the network by depth than by width.
% \jccomment{Not sure it goes here, but should we mention somewhere that balancing between old and new directions is a tradeoff that can be somewhat controlled with data sampling?}

\section{Effect of each technique}

\subsection{Ablation study} 
To understand the effect of each introduced technique, we conduct an ablation study where each model is trained on the same configuration except for one essential element (i.e., proper initialization, learning rate scaling, data up-sampling).
As a naive baseline, while having scaled learning rate and data upsampling, we include the models where no weights from the seed M20 model are used (``random init all'') to compare with a less naive baseline ``random init new'' where the M20 weights are reused and only the newly added parameters (i.e., new token embeddings in all three models, new fully-connected layer weights in \mtwide\ and weights of new layers in \mtdeep) are randomly initialized. To summarize, each configuration in Table~\ref{tab:ablation} corresponds to the following:
% \maha{Double check}
\begin{description}
\item [Random init all]: All parameters are initialized randomly while model is trained with data up-sampling and learning rate scaling.
\item[Random init new]: Newly added parameters are initialized randomly while weights of M20 are copied to the new model. Model is trained with data up-sampling and learning rate scaling.
\item[No up-sampling]: Model weights are properly initialized and their learning rates are scaled during training. No language pair is up-sampled.
\item[No lr scaling]: Model weights are properly initialized and low-resource pairs are up-sampled whereas no learning rate scaling is applied.
\end{description}

Results in Table~\ref{tab:ablation} confirm the contribution of each of the three introduced techniques in achieving the desired performance across different settings.
% impact of M20 weights
% Overall, not reusing the M20 weights leads to much slower convergence which yields the lowest spBLEU scores in all three settings. 
Overall, not reusing the M20 weights leads to worse performance than the baseline by 2$\sim$6 BLEU in different settings. 
% \maha{The table doesn't support the claim about speed of convergence, maybe word it as how much of the baseline can we recover with 30\% compute}
% impact of initialization of new weights
While reusing the old model's weights, also having proper initialization of the \emph{new} parameters yields better performance than simply initializing with default normal distribution. The benefit is most obvious when training into a wider model (29.9 vs. 32.6) compared to the other two settings.  
% \maha{can we reorder the sections and put 4.1 after 4.4 to avoid expressions like 'compared to the other two settings' without explaining what these two settings are?}
% impact of data upsampling
We also observe that data up-sampling is crucial to achieving good performance on the new directions. Not applying up-sampling degrades around $\downarrow$ 3BLEU on new directions across all settings, while barely or just slightly hurting the performance on the old directions. 
% impact of lr scaling
On the other hand, not applying learning rate scaling harms the performance of old directions across all settings, which suggests the effectiveness of scaling the learning rate to mitigate catastrophic forgetting, about which we include a more detailed analysis in section~\ref{subsec:catastrophic}.\footnote{The overall differences are small as the results are averaged over 50 directions, however, the paired T-Test shows statistically significant improvements.}
% effects of lr scale and data upsampling are additive to each other
Since learning rate scaling helps counteract catastrophic forgetting and up-sampling speeds up learning of the new directions, we discover that their effectiveness are additive -- better performance can be achieved on both old and new directions by combining these two techniques.

\begin{table}[t]
    \centering
   \scalebox{0.99}{
   \begin{tabular}{@{}lccc@{}}
\toprule
     & \multicolumn{1}{l}{\textbf{All}} & \textbf{Orig.}      & \textbf{Added}      \\ \midrule
\textbf{Mt25}       & \multicolumn{1}{l}{}       & \multicolumn{1}{l}{} & \multicolumn{1}{l}{} \\
\hspace{2em} all emb random      & 30.7   & 32.1        & 25.1        \\
\hspace{2em} new random          & 31.5   & 33.0        & 25.5        \\
\hspace{2em} new \textless{}unk\textgreater{} & \textbf{31.5}        & \textbf{33.0}     & \textbf{25.7}     \\ \midrule
\textbf{Mt25$_{wide}$}      & \multicolumn{1}{l}{}       & \multicolumn{1}{l}{} & \multicolumn{1}{l}{} \\
\hspace{2em} new random          & 31.1   & 32.3        & 26.1        \\
\hspace{2em} linear interp       & 31.8   & 33.3        & 26.0        \\
\hspace{2em} concat           & \textbf{32.0}        & \textbf{33.5}     & \textbf{26.1}     \\ \midrule
\textbf{M20t25$_{deep}$}    & \multicolumn{1}{l}{}       & \multicolumn{1}{l}{} & \multicolumn{1}{l}{} \\
\hspace{2em} random           & 31.5   & 32.8        & 26.3        \\
\hspace{2em} closest layer        & 31.6   & 32.9        & 26.3        \\
\hspace{2em} average layer      & \textbf{31.6}        & \textbf{32.9}     & \textbf{26.7}     \\ \bottomrule
\end{tabular}
   }
    \caption{Initializing the new embeddings with that of \texttt{<unk>}, concatenating the original weight with a noisy version of itself in \mtwide\ and initializing new layer in encoder/decoder with the averaged encoder/decoder weights in \mtdeep\ achieve the best performance.\tablefootnote{We did not apply learning rate scaling in this setup and do not up-sample the old related low-resource languages. All evaluations are done after 30K updates.}}
    \label{tab:initialization}
\end{table}

\subsection{Effect of proper initialization} \label{subsec:init}
In this section we briefly discuss different variations we attempted for initializing new parameters and present the results in Table~\ref{tab:initialization}. 
% vocab mismatch
\paragraph{Mt25} In the case of having only mismatched vocabularies, we find that dropping the entire old embedding table ("all emb random") causes large drop in performance, and that initializing the new embeddings with that of \texttt{<unk>} leads to slightly better performance than random initialization. 
% \jccomment{I don't totall follow that sentence. Maybe reword para?} 
\paragraph{\mtwide} A naive method to initialize wider feed-forward layer projections is to expand the old weight matrix with a randomly initialized weight matrix.
However, having random additional parameters messes up the output of feed-forward layer, which interferes with the inter-dependency among layers, thus we also tried concatenating the old weight matrix with a noisy version of itself (concat\footnote{We inject zero-mean Gaussian noise with $std=0.01$. We also tried not adding noise to the new parameters, which has almost identical performance. To avoid both parts of the weight matrices learning redundant information, we decide to add small perturbation to the new parameters.})
% \maha{Exact noising, Gaussian/uniform and params}
 followed by a normalization operation\footnote{We normalize the new weight matrix such that it has similar Frobenius norm as the old weight matrix.} 
% \maha{what normalization?}
to keep the output as close to the original projection output as possible. In addition, rather than maintaining the old weights in block, 
% \maha{do you mean in block as in contiguous?}
we also tried linearly interpolating the original weight matrices along the expanded dimension (linear interp). 
It's important that the new parameters in the two projection matrices in a feed-forward layer match each other along the hidden dimension axis.  Results indicate that weight concatenation is a simple and effective way that allows for faster convergence.

\begin{table}[!t]
    \centering
\scalebox{0.9}{
\begin{tabular}{@{}lccc@{}}
\toprule
    \textbf{Up-sampling config.}      & \textbf{All}      & \textbf{Orig.}      & \textbf{Added}      \\ \midrule
$\alpha=1$ (No up-sampling)   & 31.0        & \textbf{33.2}        & 22.2        \\
$\alpha=5$      & 31.5$^*$        & 33.0$^*$        & \textbf{25.7}$^*$        \\
$\alpha=10$       & 31.1        & 32.6$^*$        & 25.5$^*$        \\
$\alpha=5$ + rel. low-resource & \textbf{31.5}$^*$ & 33.1 & 25.1$^*$ \\ \bottomrule
\end{tabular}
}
    \caption{Performance of Mt25 at 30K updates without applying learning rate scaling. Up-sampling new languages with a reasonable ratio during continual learning leads to large gains on these new directions. $^*$ indicates $p$-value of paired T-test against the baseline (top row) is smaller than $0.05$.
    % \maha{bold the best in all tables}
    }
    \label{tab:upsampling}
\end{table}

\begin{figure}[!t]
    \includegraphics[width=0.45\textwidth]{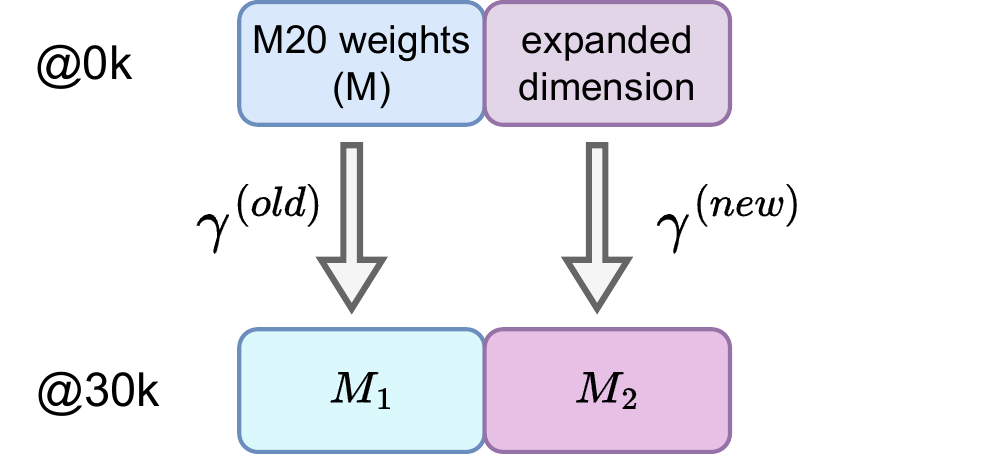}
  \caption{Before the continual learning (@0k), the wider hidden projection matrix is initialized with a concatenation of M20 weights $M$ and a noisy version of itself. During the continual learning stage, the learning rate for weights copied from old model is scaled by $\gamma^{(old)}$, the rest are scaled by $\gamma^{(new)}$.  }
    \label{fig:lr_scale}
\end{figure}

% deeper model
\paragraph{\mtdeep}  Our preliminary experiments show that inserting new layers at the bottom of the encoder and at the top of the decoder is more effective than other attempted scenarios, so we went with this setup in all our \mtdeep\ experiments. We tried different methods of initializing the inserted layers, including random initialization, copying parameters from the closest layer,\footnote{Since we include the new layer at the bottom of the encoder and at the top of the decoder, the closest layer means the first encoder layer and last decoder layer respectively.} 
% \maha{I'd go for putting the footnotes in the main text}
% \jccomment{"Closest"? This is in different "directions" depending for encoder and decoder, right?}
and averaging the weights across all encoder or decoder layers.  Although the benefit is small, we do perceive that initializing the new layers with averaged layer weights results in better performance, especially on new translation directions (26.7 vs. 26.3). 
% \jccomment{Separate averages for encoder and decoder?}

\subsection{Effect of data up-sampling} \label{subsec:upsample}
We multiply the original dataset size of each up-sampled direction by a value $\alpha$ before computing the final up-sample ratio.
The new directions thus constitute a larger portion in the new dataset than if not. We show in Table~\ref{tab:upsampling} the effect of the up-sample factor $\alpha$. We find that up-sampling selected languages leads to much better performance (+3BLEU) on new directions than without up-sampling.  However, up-sampling the new directions too much (e.g., $\alpha=10$) worsens the performance on old directions while not improving the new directions. In general, we find the up-sampling factor $\alpha=5$ adopted in previous studies~\cite{garcia-etal-2021-towards,berard-2021-continual}  suitable to many other variants we have attempted in this work.
% \jccomment{To discuss: as I mentioned in Sect 2, needing a hyperparameter search undermines the claim of computational efficiency} \sscomment{agreed, but haven't thought of a better way to describe this part. I think it's also true for learning rate scaling, we need to tune the scaling factor for lr scaling.}
Besides merely up-sampling the new directions, we also up-sample the old low-resource directions that belong to the same language family as any of the new directions. Doing so slightly improves the performance on old directions whereas causes a drop in new directions, which further confirms the critical role of up-sampling in balancing the performance between the old and new directions.
% \maha{on 'All' it's the same score and the drop in the new is more significant than the gain on Orig.! - the claim of  "confirms the effectiveness of data up-sampling" is more related to the comparison between no up-sampling (row 1) and alpha=5 (row 2) than to the comparison between rows 2 and 4}

\begin{table}[t]
    \centering
\scalebox{0.87}{
\begin{tabular}{@{}lccc@{}}
\toprule
        & \multicolumn{1}{l}{\textbf{All}} & \textbf{Orig.} & \multicolumn{1}{l}{\textbf{Added}} \\ \midrule
        $\gamma^{(old)} = 1$,  \hspace{0.61em} $\gamma^{(new)} = 1$    & 32.0   & 33.5      & 26.1   \\
$\gamma^{(old)} = 0.5$,  $\gamma^{(new)} = 5$ \hspace{1em} & \bf 32.3$^*$   & \bf 33.8$^*$      & \bf 26.2   \\
$\gamma^{(old)} = 0.5$,  $\gamma^{(new)} = 0.5$    & 31.5$^*$   & 33.3      & 24.7$^*$   \\ 
$\gamma^{(old)} = 0.5$,  $\gamma^{(new)} = 1$    & 32.1$^*$   & 33.6$^*$      & 26.0   \\
$\gamma^{(old)} = 5$, \hspace{0.6em} $\gamma^{(new)} = 5$      & 31.9   & 33.4      & 26.1   \\ \bottomrule
\end{tabular}
}
    \caption{Performance of \mtwide\ after 30K updates while applying different learning rate scaling factor based on notation in Figure~\ref{fig:lr_scale}. These experiments adopt an earlier setup where the old related low-resource languages are not up-sampled. $^*$ indicates $p$-value of paired T-test against the baseline (top row) is smaller than $0.05$.}
    \label{tab:lr_scaling}
\end{table}

\subsection{Effect of learning rate scaling} \label{subsec:lr_scale}
% general idea of learning rate scaling, refer to the fig
As described in section~\ref{sec:method}, we scale down the learning rate for old (reused) parameters. In the case of Mt25, all parameters are updated with a smaller learning rate than the new token embeddings. In the wider network, as we have established the effectiveness of using concatenated weights, we apply different learning rate to each part of the weight matrices as generically illustrated in Figure~\ref{fig:lr_scale}.
In \mtdeep, the new layers are updated with a larger learning rate while the rest of the parameters receive a smaller learning rate. 
% \maha{I don't like the term speed. A higher learning rate doesn't mean fast learning}
% \jccomment{This is for deeper nets right?}
% scaling the update speed of different part of the network leads to better performance
Table~\ref{tab:lr_scaling} suggests that having a smaller learning rate for old parameters is more favorable than scaling all parameters by the same amount ratio. Scaling down the update for all parameters slows down the learning of the new directions, whereas scaling up with a larger value does not improve the performance either. While already applying smaller update on old parameters, scaling up the learning rate for the new parameters can in fact improve both the old and new directions (the top two rows in Table~\ref{tab:lr_scaling}). Overall we find that learning rate scaling is an effective and easy-to-implement alternative to previous methods~\citep{EWC} in terms of alleviating catastrophic forgetting. 
% \maha{Can you add details about the learning rate schedule during the two phases (I believe you do not reset the lr scheduler) so that we get a sense of the magnitude of the lr rate in the 2nd phase}

\section{Analysis}
% \maha{a few sentences to motivate/outline}

\begin{table}[t]
    \centering
\scalebox{0.9}{
\begin{tabular}{@{}lcccc@{}}
\toprule
\multicolumn{1}{l}{} & \textbf{Mt25} & \textbf{Mt25\_v1} & \textbf{Mt25\_v2} & \textbf{M12t25} \\ \midrule
All     & 32.6  & 32.7      & 32.5  & 32.7    \\
High    & 35.8  & 36.0      & 35.8   & 35.9   \\
Mid     & 29.6  & 29.8      & 29.7   & 29.3   \\
Low     & 31.5  & 31.2      & 31.4  & 31.6    \\
V\_Low  & 19.0  & 20.0      & 19.0   & 20.1   \\ \bottomrule
\end{tabular}
}
    \caption{Besides the language breakdown shown in Table~\ref{tab:langbreakdown}, we show our approach also generalizes to other groupings of old and new languages (Mt25\_v1, Mt25\_v2 and M12t25), details about which are included in Appendix~\ref{sec:language_selection} Table~\ref{tab:lang_breakdown_new}. }
    \label{tab:seed_languages}
\end{table}

\begin{table}[t]
    \centering
    \scalebox{0.8}{
    \begin{tabular}{@{}ccc|cc|cc@{}}
\toprule
\multicolumn{1}{l}{} & \multicolumn{2}{c|}{\textbf{Mt25}} & \multicolumn{2}{c|}{\textbf{Mt25\_v1}} & \multicolumn{2}{c}{\textbf{Mt25\_v2}} \\ \cmidrule(l){2-7} 
\multicolumn{1}{l}{} & $\Delta$BLEU   & Up   & $\Delta$BLEU   & Up   & $\Delta$BLEU   & Up  \\ \midrule
eng-bel      & -1.2     & N    & 2.0      & Y    & 2.4      & Y           \\
eng-guj      & 1.6      & Y    & 0.9      & Y    & 0.0      & N           \\
eng-npi      & 0.8      & Y    & -0.1     & Y    & -2.3     & N           \\
eng-xho      & 1.3      & Y    & 0.0      & N    & 1.1      & Y           \\ \midrule
bel-eng      & -1.1     & N    & -0.5     & Y    & -0.7     & Y           \\
guj-eng      & -2.5     & Y    & -2.6     & Y    & -2.2     & N           \\
npi-eng      & -1.6     & Y    & -0.6     & Y    & -1.9     & N           \\
xho-eng      & -1.1     & Y    & -1.3     & N    & -0.7     & Y           \\ \bottomrule
\end{tabular}
    }
    \caption{Zoomed in analysis over specific language pairs. Our approach is more effective in learning eng$\rightarrow$xxx directions than xxx$\rightarrow$eng directions across different seed-language setups regardless if the language is up-sampled (Up=Y) or not (Up=N).}
    \label{tab:seed_lang_fine_grained}
\end{table}

\subsection{How does the choice of seed languages affect continual learning?} \label{subsec:seed_lang}

In addition to the core setup described in Table~\ref{tab:langbreakdown}, we also experiment with other settings.
% Besides the seed 20 languages described in the previous experimental setup, we also experiment with training the seed M20 model on other sets of initial languages. We include more details on these language groupings in Appendix~\ref{sec:language_selection}. 
% Since it is computationally infeasible to try out all of the old/new language combinations,\maha{Rephrase, this is obvious and our goal isn't to sweep but to experiment with different scenarios} 
We considered three new setups following the natural scenario where the seed languages are mostly of high- and mid-resource languages. Two of the new setups still adopt the 20/5 division between old and new languages, but emphasize different scenarios --- Mt25\_v1 covers one mid resource language in the set of new languages and Mt25\_v2 does not add any new script. The last one instead initializes from a model trained on 12 high- and mid-resource languages and adds in 13 new languages.\footnote{For detailed breakdown of these three new settings, we refer readers to Table~\ref{tab:lang_breakdown_new} and Table~\ref{tab:m12t25} in Appendix~\ref{sec:language_selection}}
We follow the same configuration used in the previous experiments and train into wider models in the continual learning phase. Results in Table~\ref{tab:seed_languages} demonstrate that our approach successfully generalizes to all the new settings, as similar performances can be achieved when models are trained on different choices of seed languages, even when the number of seed languages are different. However, we do observe small differences after zooming in by resource-levels. A more fine-grained analysis over a few low-resource languages that cover diverse scripts (bel-Cyrl, guj-Gujr, npi-Deva, xho-Latn) is presented in Table~\ref{tab:seed_lang_fine_grained}. In general, regardless of if the language pair is in the seed language set or not, the eng$\rightarrow$xxx directions are learned faster than xxx$\rightarrow$eng directions, since most of the pairs already exceed the baseline performance after just 30K updates (compare the upper part of Table~\ref{tab:seed_lang_fine_grained} to the lower part). We also verify the effectiveness of data up-sampling for eng$\rightarrow$xxx directions. Up-sampling these pairs leads to much better performance than when not up-sampling (e.g., compare performance of eng-bel in the three setups).

\begin{figure}[!t]
    \centering
    \includegraphics[width=0.5\textwidth]{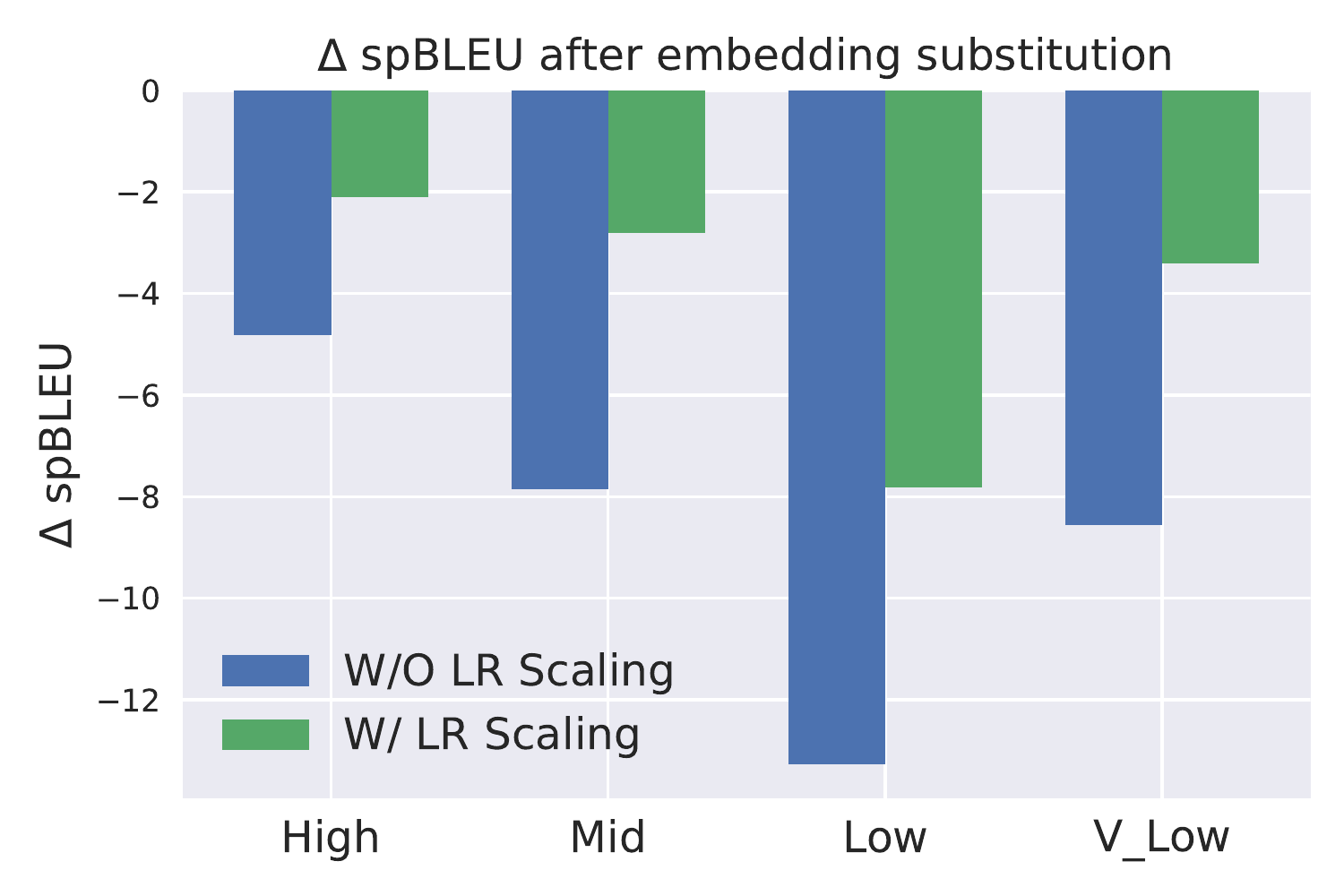}
    \caption{We measure the drop in spBLEU on original directions after substituting overlapped embeddings in Mt25$_{wide}$ back to M20 model. Applying learning rate scaling helps alleviate catastrophic forgetting as spBLEU drops less than without LR scaling. }
    \label{fig:bleu_drop}
\end{figure}

\subsection{How effective is learning rate scaling in
mitigating catastrophic forgetting?} \label{subsec:catastrophic}
To quantitatively measure the amount of information lost after the continual learning phase, we adopt an evaluation setup akin to ~\cite{garcia-etal-2021-towards}, in which the embeddings in M20 that overlap with Mt25 are substituted with the corresponding embeddings in Mt25. One can evaluate this new M20 model with substituted embeddings on the original 20 languages, and use the drop in spBLEU as a proxy for the amount of knowledge lost in the embeddings due to catastrophic forgetting. In Figure~\ref{fig:bleu_drop}, we display the spBLEU drop after substituting the embeddings of a \mtwide\ model trained with and without learning rate scaling. For both variants, the spBLEU scores drop, indicating that some information in the embeddings is lost after training on the new languages. We also find that, for both models, the decrease in spBLEU is larger for (very) low-resource languages than high and mid resource languages, which suggests (very) low-resource languages are easier to be ``forgotten'' in the second training phase, which reinforces our decision of also up-sampling the related low-resource languages as introduced in section~\ref{subsec:upsample}. Finally, not applying learning rate scaling leads to much larger decrease in all directions, which manifests the effectiveness of scaling down the learning rate for alleviating catastrophic forgetting.\footnote{Besides probing the effect of learning rate scaling on embeddings, we also present an analysis on the expanded feed-forward weights in Appendix~\ref{sec:norm_analysis}.}
% \maha{either go with extremely low or very low and use the same name throughout the paper}

\begin{figure}[!t]
    \centering
    \includegraphics[width=0.45\textwidth]{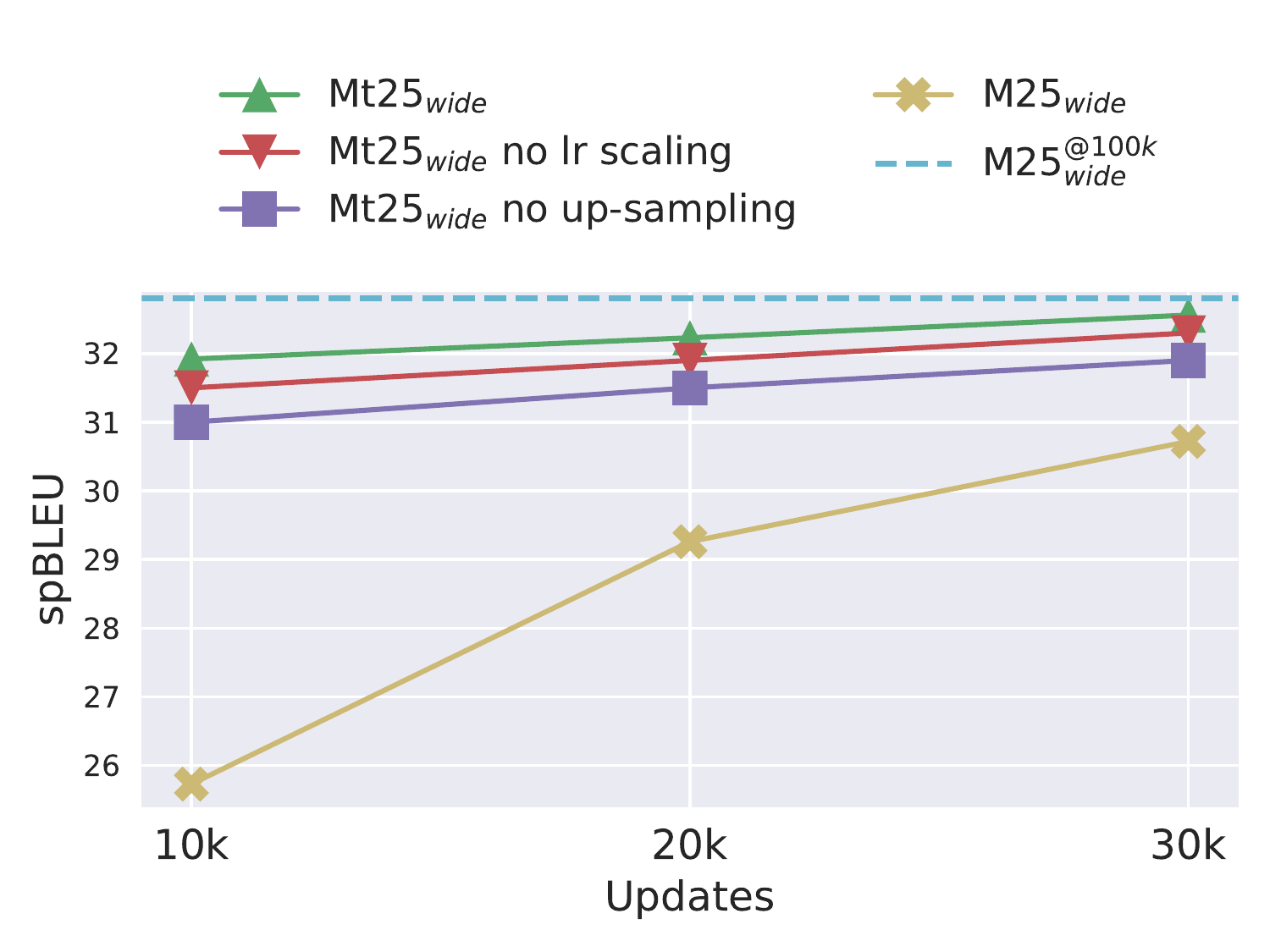}
    \caption{Training the M25 from scratch without using the M20 weights converges significantly slower. Applying the techniques introduced in this work results in the largest computation reduction, achieving over 95\% performance with less than 10\% baseline computation.}
    \label{fig:spbleu_vs_updates}
\end{figure}

\subsection{Computation saved}
Due to the mismatched vocabularies and architectures, models trained on the combination of new and old languages are typically re-trained from scratch after already incurring large computation on old languages. In this section, we look at how much computation can be effectively saved with the approach proposed in this paper. Note that, for both our methods and the retraining approach, M20 is already trained, and its computation cost is excluded from our calculation.  

Although results in Table~\ref{tab:main_res} shows that 50\% of the baseline computation is required to recover the baseline performance, we show in Figure~\ref{fig:spbleu_vs_updates} that much computation can be saved if we slightly slack the target performance. The \mwide\ model trained from scratch reaches only $78\%$ of the baseline performance after 10k updates, while models trained with the combination of our techniques achieve over $97\%$ of the baseline performance after the same amount of updates (10\% total computation of training \mtwide).

\section{Related Work}

% adding new languages to multilingual machine translation models
Our work is closely related to prior research on adapting existing MMT models to new languages~\citep{lakew-etal-2019-adapting,Kocmi2020ExploringBO}. \citet{neubig-hu-2018-rapid} add low-resource languages to multilingual models by fine-tuning on low-resource data while regularizing with related high-resource data. ~\citet{garcia-etal-2021-towards} introduce simple vocabulary substitution for adapting MMT models to new languages without any architectural changes.  Another line of research employs modular approaches, which include training lightweight adapters~\citep{bapna-firat-2019-simple}, language-specific encoder-decoders~\citep{escolano-etal-2019-bilingual,escolano-etal-2021-multilingual}, language specific embeddings~\citep{berard-2021-continual} for learning new languages. While sometimes escaping the need to train on old examples, growing the model in a modular fashion~\citep{progressive_net} requires non-trivial changes to standard architectures. In contrast, our work relies on rehearsal mechanism (i.e., also train on old examples) but does not need to modify network structures~\citep{robins1995catastrophic}. 
% common approaches to alleviating catastrophic forgetting in continual learning + several works on growing architectures

Our approach is also related to works that focus on continual learning of MT models for adapting multiple domains. \citet{thompson-etal-2019-overcoming,gu-feng-2020-investigating} adopt a method derived from Elastic Weight Consolidation~\citep{EWC} to alleviate catastrophic forgetting. While most prior works only investigate two-stage continual learning, \citet{cao-etal-2021-continual} propose a new framework that extends to multi-stage training to mitigate catastrophic forgetting~\citep{ring1994continual}. The initialization of the new parameters and embeddings in our technique is also related to that in \citep{pfeiffer-etal-2021-unks}, which accommodate multilingual models to unseen scripts via matrix factorization. Our focus on the architectural differences between initial and continual learning phase is also relevant to recent discoveries that wider networks forget less catastrophically~\citep{mirzadeh2022architecture}.

\section{Conclusion}

We show in this work that it is possible to efficiently bootstrap from existing models and recover the baseline performance  with much less computation while vocabularies and architectures can be different in the continue learning stage. We highlight the importance of (1) reusing the existing model weights and carefully initializing the new parameters, (2) applying learning rate scaling, and (3) performing data up-sampling. Analyses reveal that scaling down the learning rate for old parameters helps alleviate catastrophic forgetting, and that data up-sampling is vital to achieving good performance on the new directions. We hope our work can help save computation for research into large-scale multilingual MT models, and more generally, will help spur research into continual multitask learning in the presence of architectural changes.
\section*{Limitations}

While we explore the under-studied architectural mismatches for continual learning of MMT models, we focus exclusively on adding new languages in bulk, without investigating adding languages one by one continuously. Furthermore, due to limited computational resources, we only experimented with a few typical scenarios where the new languages are low or very-low resourced. Experiments on other groupings of old and new languages could further validate the effectiveness of our approach.

\section*{Acknowledgement}

We thank Shruti Bhosale, Jean Maillard, Philipp Koehn, Francisco Guzm\'{a}n, Angela Fan, Holger Schwenk, Haoran Xu, Steven Tan and Alex Guo for their helpful discussion and support. We thank Vedanuj Goswami for helping fix technical issue at the end of this project.

% \section*{Acknowledgements}

\bibliography{anthology,custom}
\bibliographystyle{acl_natbib}

\newpage

\appendix
\section{Language details}
\label{sec:language_selection}

% \begin{table}[ht]
%     \centering
%     \begin{tabular}{@{}cccc|c@{}}
% \toprule
% \multicolumn{4}{c|}{\textbf{Original}} & \textbf{Added} \\ \midrule
% lav   & rus   & fin   & spa   & xho   \\
% lit   & hin   & est   & deu   & guj   \\
% swh   & mar   & pol   & zul$^*$   & npi   \\
% ukr   & mkd   & ces   & msa$^*$   & ind   \\
% bel   & bul   & fra   & kir$^*$   & kaz   \\ \bottomrule
% \end{tabular}
%     \caption{The M20 model is trained on the ``Original'' languages. Mt25, \mtwide\ and \mtdeep\ bootstrap from M20 and train on the combination of ``Original'' and ``Added'' languages. We perform data up-sampling over added data in conjunction with related old low-resource languages (marked with $^*$).
%     }
%     \label{tab:langbreakdown}
% \end{table}

\begin{figure}[!t]
    \includegraphics[width=0.5\textwidth]{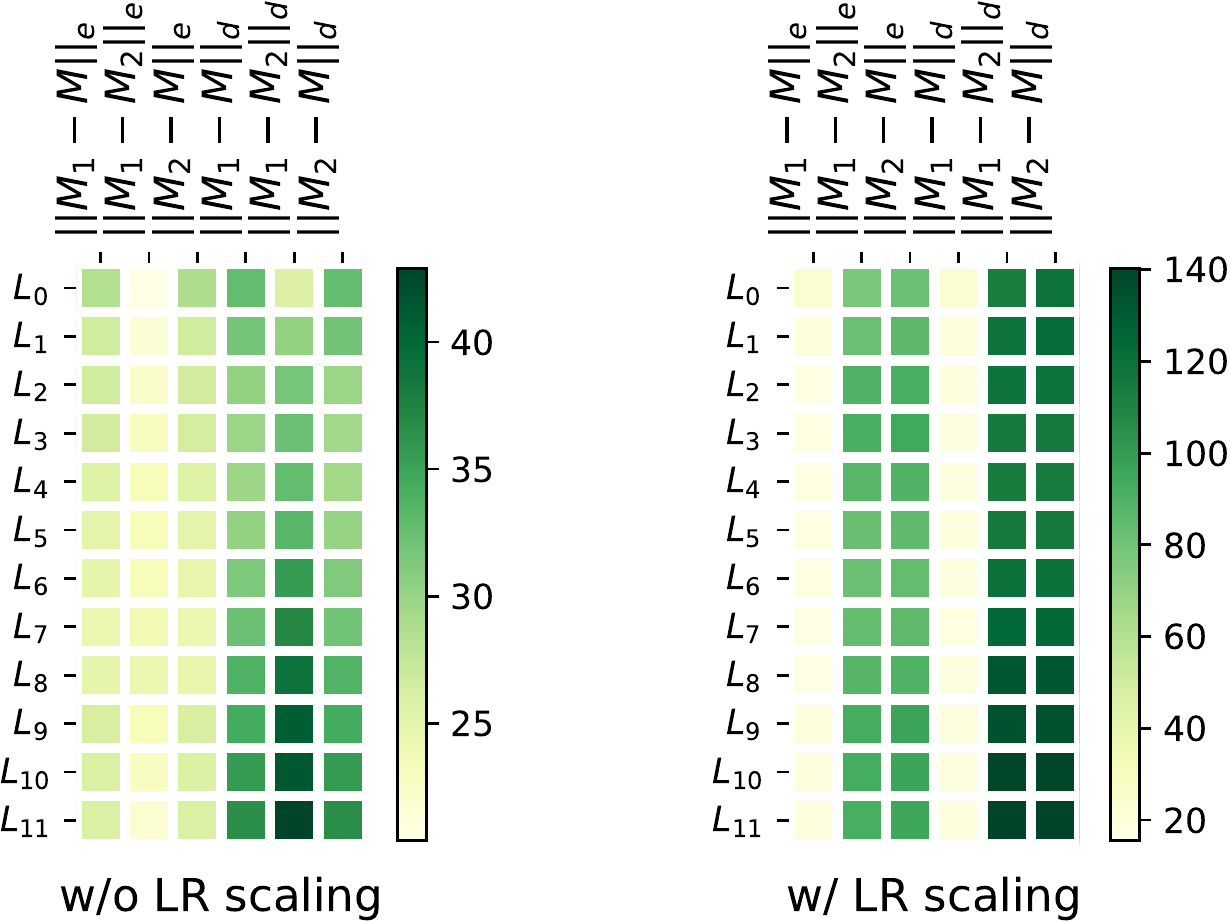}
    \caption{ We measure the amount of change in each feed-forward layer's projection weights via Frobenius norm ($\|\cdot\|_e$ for encoder layers and $\|\cdot\|_d$ for decoder layers).  Notation in this figure follows that in Figure~\ref{fig:lr_scale}. Scaling down the learning rate for parameters copied from the old model keeps them close to their initialization while pushes the newly added parameters farther away.
    }
    \label{fig:fro_norm}
\end{figure}

We present two extra language grouping settings we have experimented in this section. To verify the validity of our approach on other language groupings, we also experiment with two other settings as shown in Table~\ref{tab:lang_breakdown_new}. In addition to dividing the old/new languages by 20/5, we also tried a setting where the seed model is trained on fewer languages as reported in Table~\ref{tab:m12t25}.
\begin{table*}[ht]
    \centering
    %\scalebox{0.9}{
    \resizebox{\linewidth}{!}{%
    \begin{tabular}{@{}cllllll@{}}
\toprule
\multicolumn{1}{l}{}                         & \multicolumn{1}{l}{\textbf{Language}} & \multicolumn{1}{l}{\textbf{Language code}} &  \multicolumn{1}{l}{\textbf{Language family}} & \textbf{\# bitext} & \textbf{Res. level} & \multicolumn{1}{l}{\textbf{Script}} \\ \midrule
\multirow{20}{*}{\textbf{M20 old languages}} & Latvian                               & lav                                        & Baltic                                       & 2,288,307          & Mid                 & Latn                                \\
                                             & Lithuanian                            & lit                                        & Baltic                                       & 2,322,619          & Mid                 & Latn                                \\
                                             & Swahili                               & swh                                        & Benue-Congo                                  & 1,133,371          & Mid                 & Latn                                \\
                                             & Zulu                                  & zul                                        & Benue-Congo                                  & 656,450            & Low                 & Latn                                \\
                                             & Ukrainian                             & ukr                                        & East Slavic                                  & 925,922            & Low                 & Cyrl                                \\
                                             & Belarusian                            & bel                                        & East Slavic                                  & 55,185             & V\_Low              & Cyrl                                \\
                                             & Russian                               & rus                                        & East Slavic                                  & 13,962,768         & High                & Cyrl                                \\
                                             & Hindi                                 & hin                                        & Indo-Iranian                                 & 1,032,502          & Mid                 & Deva                                \\
                                             & Marathi                               & mar                                        & Indo-Iranian                                 & 181,642            & Low                 & Deva                                \\
                                             & Malay                                 & msa                                        & Malayo-Polynesian                            & 305,469            & Low                 & Latn                                \\
                                             & Macedonian                            & mkd                                        & South Slavic                                 & 403,742            & Low                 & Cyrl                                \\
                                             & Bulgarian                             & bul                                        & South Slavic                                 & 2,118,839          & Mid                 & Cyrl                                \\
                                             & Kyrgyz                                & kir                                        & Turkic                                       & 336,554            & Low                 & Cyrl                                \\
                                             & Finnish                               & fin                                        & Uralic                                       & 3,764,770          & Mid                 & Latn                                \\
                                             & Estonian                              & est                                        & Uralic                                       & 2,410,543          & Mid                 & Latn                                \\
                                             & Polish                                & pol                                        & West Slavic                                  & 3,062,818          & Mid                 & Latn                                \\
                                             & Czech                                 & ces                                        & West Slavic                                  & 23,792,604         & High                & Latn                                \\
                                             & French                                & fra                                        & Western European                             & 20,031,051         & High                & Latn                                \\
                                             & Spanish                               & spa                                        & Western European                             & 16,606,594         & High                & Latn                                \\
                                             & German                                & deu                                        & Western European                             & 10,198,897         & High                & Latn                                \\ \midrule
\multirow{5}{*}{\textbf{Mt25 new languages}} & Xhosa                                 & xho                                        & Benue-Congo                                  & 634,078            & Low                 & Latn                                \\
                                             & Gujarati                              & guj                                        & Indo-Iranian                                 & 153,985            & Low                 & Gujr                                \\
                                             & Nepali                                & npi                                        & Indo-Iranian                                 & 72,250             & V\_Low              & Deva                                \\
                                             & Indonesian                            & ind                                        & Malayo-Polynesian                            & 771,801            & Low                 & Latn                                \\
                                             & Kazakh                                & kaz                                        & Turkic                                       & 627,734            & Low                 & Cyrl                                \\ \bottomrule
\end{tabular}
    }
    \caption{Detailed information about languages used in the main setup of our experiments. Languages having >10M examples are of high-resource languages, having (1M, 10M] mid-resource languages. The rest are of low-resource languages, v\_low is a subset of low-resource languages that have <100K examples.}
    \label{tab:language_details}
\end{table*}
%\FloatBarrier

\begin{table*}[ht]
    \centering
    \begin{tabular}{@{}cccc|c@{}}
\toprule
\multicolumn{4}{c|}{\textbf{Mt25\_v1 Original}} & \multicolumn{1}{l}{\textbf{Added}} \\ \midrule
lav     & ukr$^*$     & msa      & pol    & guj                              \\
lit     & rus      & mkd      & ces    & npi                              \\
swh     & hin      & bul      & fra    & est                              \\
zul     & mar$^*$     & fin      & spa    & bel                              \\
xho     & ind      & kaz$^*$     & deu    & kir                              \\ \bottomrule
\end{tabular} \hspace{1em} \begin{tabular}{@{}cccc|c@{}}
\toprule
\multicolumn{4}{c|}{\textbf{Mt25\_v2 Original}} & \multicolumn{1}{l}{\textbf{Added}} \\ \midrule
lav      & hin      & mkd     & pol    & zul                              \\
lit      & mar      & bul     & ces    & xho                              \\
swh      & guj      & kaz$^*$    & fra    & ind                              \\
ukr$^*$     & npi      & fin     & spa    & bel                              \\
rus      & msa$^*$     & est     & deu    & kir                              \\ \bottomrule
\end{tabular}
    \caption{Besides the breakdown between old and new languages as shown in Table~\ref{tab:langbreakdown}, Mt25\_v1 uses the left grouping and Mt25\_v2 uses the right grouping of languages. $^*$ marks the languages that are also up-sampled during continual learning.}
    \label{tab:lang_breakdown_new}
\end{table*}

\begin{table*}[ht]
    \centering
    \begin{tabular}{@{}ll|ll@{}}
\toprule
\multicolumn{2}{c|}{\textbf{Original}} & \multicolumn{2}{c}{\textbf{Added}} \\ \midrule
lav                & est               & zul             & mar            \\
lit                & pol               & xho             & guj            \\
swh                & ces               & ukr             & npi            \\
rus                & fra               & bel             & ind            \\
bul                & spa               & hin             & msa            \\
fin                & deu               & mkd             & kaz            \\
                   &                   & kir             &                \\ \bottomrule
\end{tabular}\hspace{2em}\begin{tabular}{@{}lccc@{}}
\toprule
               & \textbf{All} & \textbf{Orig.} & \textbf{Added} \\ \midrule
f=1e-6, $\gamma^{(old)}$=0.25 & 32.3         & 33.8           & 26.4           \\
f=1e-6, $\gamma^{(old)}$=0.1  & 32.1         & 33.7           & 26.0           \\
f=1e-5, $\gamma^{(old)}$=0.25 & 31.9         & 33.3           & 26.2           \\
f=1e-5, $\gamma^{(old)}$=0.1  & 32.1         & 33.5           & 26.6           \\
f=1e-5, $\gamma^{(old)}$=0.05 & 31.8         & 33.3           & 25.8           \\
f=1e-4, $\gamma^{(old)}$=0.25 & 31.7         & 33.1           & 26.1           \\ \bottomrule
\end{tabular}
    \caption{\textbf{Left:} M12t25 starts from a model trained on 12 high- and mid- resource languages and grows to a wider model to support 13 new low- and very-low- resource languages. None of old languages is up-sampled in the continual learning stage. \textbf{Right:} Performance of \mtwide\ after 30k updates when the learning rate for parameter whose Fisher information is greater than threshold $f$ is scaled down by the corresponding $\gamma^{(old)}$ in each row. }
    \label{tab:m12t25}
\end{table*}

\begin{figure*}[ht]
    \centering
    \includegraphics[width=0.5\textwidth]{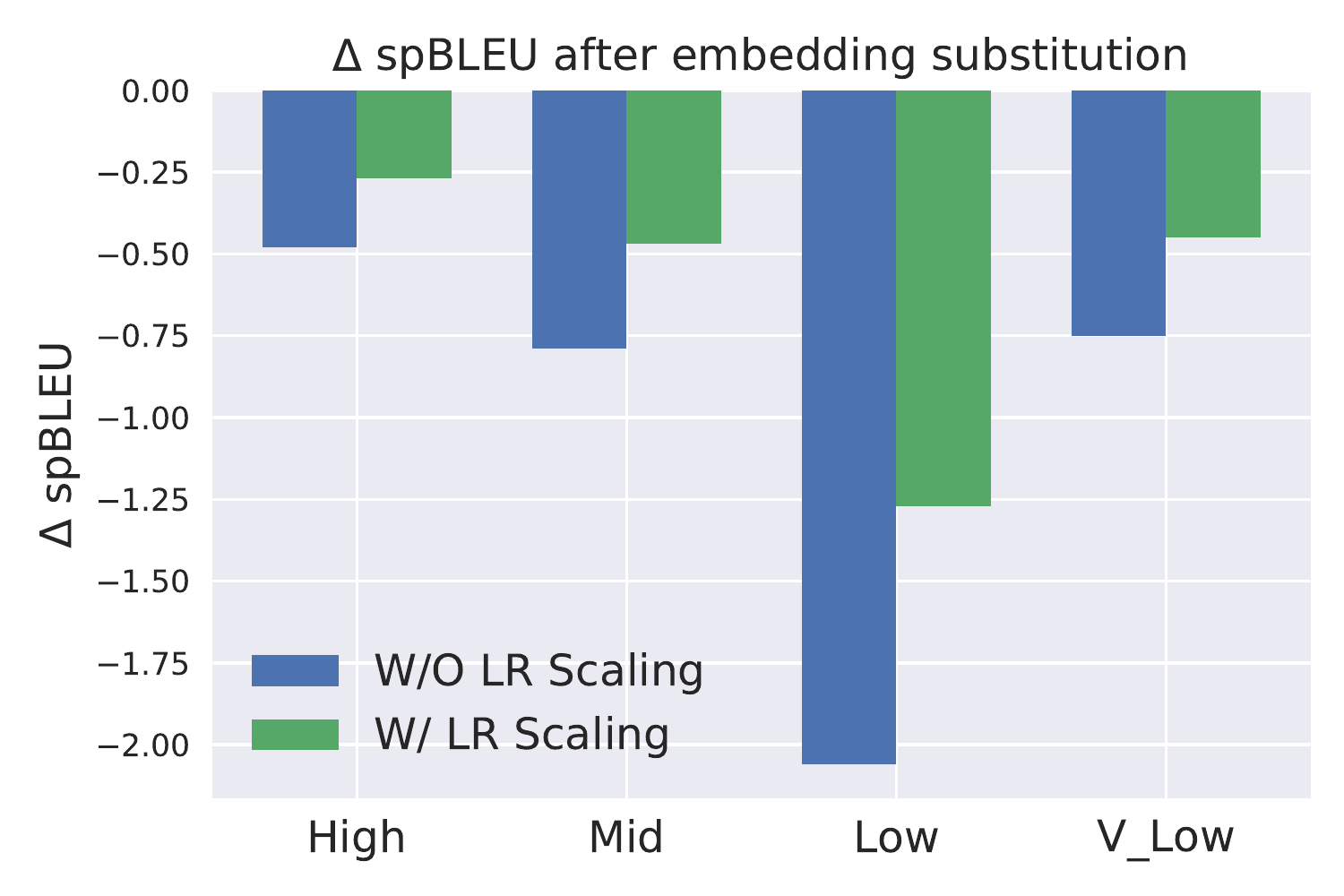}\includegraphics[width=0.5\linewidth]{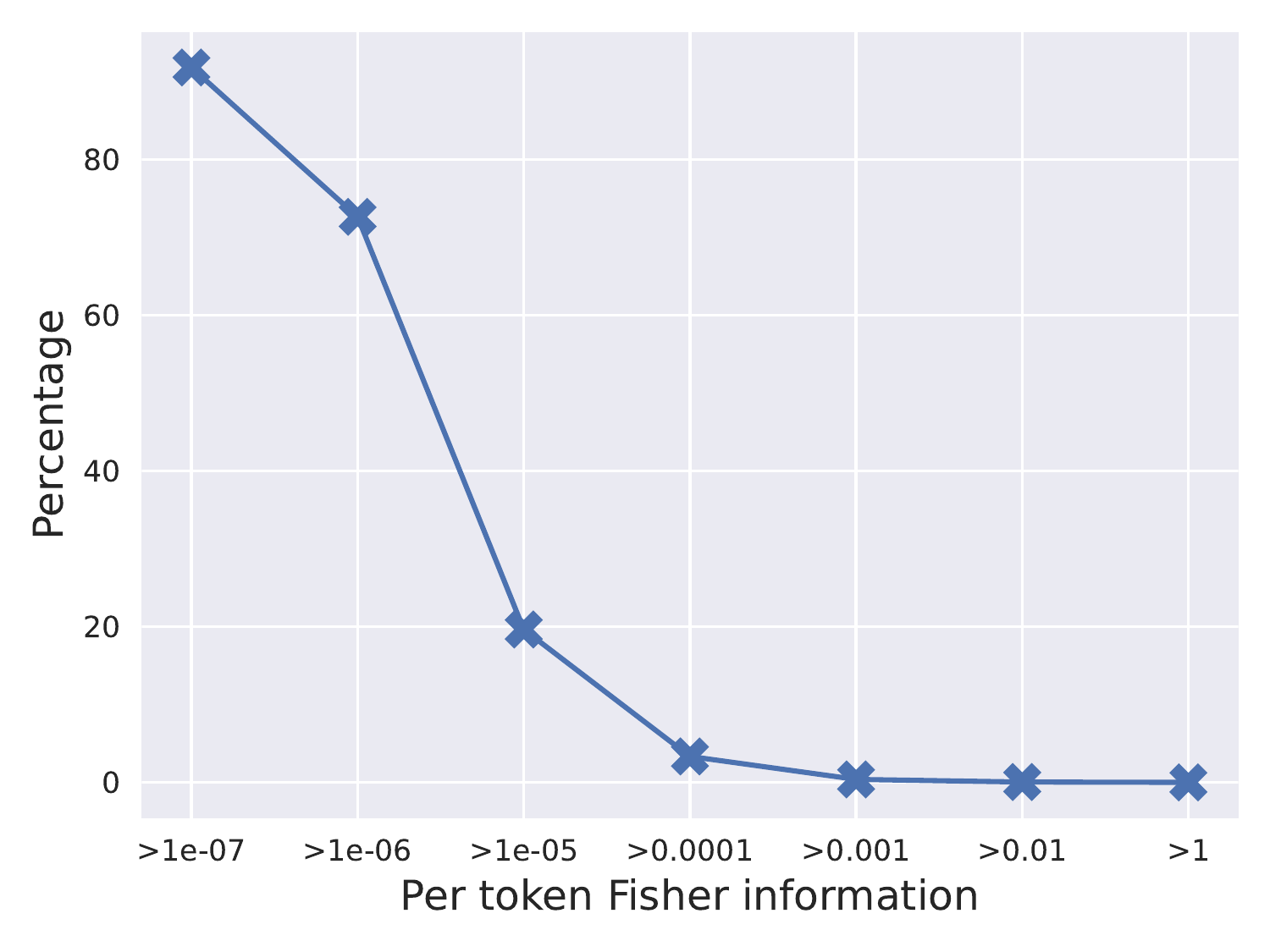}
    \caption{\textbf{Left:} spBLEU drop after substituting the embedding of the learned Mt25 model back to M20 model. \textbf{Right:} Per token Fisher information calculated over FloRes dev set. }
    \label{fig:bleu_drop_mt25}
\end{figure*}

\section{Detailed performance of each direction}

\begin{table*}[ht]
    %\scalebox{0.8}{
    \resizebox{\linewidth}{!}{%
    \begin{tabular}{@{}lcc|cc|cc|cc|cc|cc@{}}
\toprule
\textbf{Lang\_pair} & \multicolumn{2}{c}{\textbf{M25}} & \multicolumn{2}{c}{\textbf{Mt25}} & \multicolumn{2}{c}{\textbf{\mwide}} & \multicolumn{2}{c}{\textbf{Mt25$_{wide}^{(@30k)}$}} & \multicolumn{2}{c}{\textbf{\mdeep}} & \multicolumn{2}{c}{\textbf{Mt25$_{deep}^{(@30k)}$}} \\ \cmidrule(l){2-13} 
                    & \textbf{spBLEU}  & \textbf{chrF++} & \textbf{spBLEU}  & \textbf{chrF++}  & \textbf{spBLEU}     & \textbf{chrF++}    & \textbf{spBLEU}     & \textbf{chrF++}     & \textbf{spBLEU}     & \textbf{chrF++}    & \textbf{spBLEU}     & \textbf{chrF++}     \\ \midrule
bel-eng             & 18.1             & 0.74          & 18.3             & 0.74           & 20.1                & 0.74             & 19.0                & 0.74              & 19.9                & 0.74             & 18.5                & 0.74              \\
bul-eng             & 40.8             & 0.81          & 41.1             & 0.81           & 41.7                & 0.81             & 41.5                & 0.81              & 41.9                & 0.81             & 41.0                & 0.81              \\
ces-eng             & 42.6             & 0.81          & 42.3             & 0.81           & 42.9                & 0.82             & 42.5                & 0.81              & 43.0                & 0.81             & 42.0                & 0.81              \\
deu-eng             & 44.7             & 0.82          & 44.5             & 0.82           & 45.3                & 0.82             & 45.2                & 0.82              & 45.4                & 0.82             & 44.5                & 0.82              \\
eng-bel             & 11.7             & 0.59          & 11.3             & 0.59           & 13.2                & 0.61             & 12.1                & 0.59              & 13.1                & 0.60             & 12.1                & 0.60              \\
eng-bul             & 39.5             & 0.78          & 40.3             & 0.79           & 40.5                & 0.79             & 40.1                & 0.78              & 40.6                & 0.79             & 40.3                & 0.79              \\
eng-ces             & 38.4             & 0.76          & 38.5             & 0.76           & 38.4                & 0.76             & 38.2                & 0.76              & 38.8                & 0.76             & 38.4                & 0.76              \\
eng-deu             & 39.0             & 0.79          & 39.3             & 0.79           & 40.1                & 0.79             & 39.2                & 0.78              & 39.8                & 0.79             & 39.3                & 0.79              \\
eng-est             & 30.3             & 0.76          & 30.1             & 0.76           & 32.0                & 0.77             & 30.7                & 0.76              & 31.5                & 0.76             & 30.7                & 0.76              \\
eng-fin             & 28.5             & 0.76          & 28.5             & 0.76           & 30.3                & 0.76             & 29.3                & 0.76              & 30.7                & 0.77             & 29.5                & 0.76              \\
eng-fra             & 49.4             & 0.81          & 49.4             & 0.81           & 48.6                & 0.81             & 49.3                & 0.81              & 49.1                & 0.81             & 49.1                & 0.81              \\
eng-guj             & 24.2             & 0.63          & 28.3             & 0.67           & 25.5                & 0.64             & 28.4                & 0.67              & 25.3                & 0.64             & 28.0                & 0.66              \\
eng-hin             & 32.0             & 0.71          & 31.5             & 0.71           & 33.4                & 0.72             & 32.6                & 0.71              & 32.7                & 0.71             & 32.8                & 0.71              \\
eng-ind             & 40.1             & 0.81          & 40.0             & 0.81           & 41.2                & 0.81             & 44.4                & 0.82              & 41.4                & 0.81             & 43.7                & 0.82              \\
eng-kaz             & 14.6             & 0.58          & 13.4             & 0.57           & 14.8                & 0.58             & 18.0                & 0.61              & 16.4                & 0.60             & 15.6                & 0.59              \\
eng-kir             & 18.0             & 0.61          & 16.4             & 0.59           & 19.1                & 0.62             & 20.9                & 0.63              & 18.5                & 0.62             & 20.4                & 0.63              \\
eng-lav             & 32.4             & 0.75          & 33.5             & 0.75           & 33.6                & 0.75             & 34.0                & 0.75              & 34.1                & 0.75             & 33.6                & 0.75              \\
eng-lit             & 28.3             & 0.73          & 28.1             & 0.73           & 29.5                & 0.74             & 28.9                & 0.74              & 29.7                & 0.74             & 28.6                & 0.73              \\
eng-mar             & 16.9             & 0.62          & 16.3             & 0.62           & 18.2                & 0.62             & 17.8                & 0.63              & 18.5                & 0.63             & 17.3                & 0.62              \\
eng-mkd             & 34.9             & 0.77          & 35.2             & 0.77           & 36.5                & 0.77             & 35.8                & 0.77              & 35.8                & 0.77             & 35.2                & 0.77              \\
eng-msa             & 37.2             & 0.80          & 32.5             & 0.79           & 39.6                & 0.81             & 43.3                & 0.82              & 39.3                & 0.80             & 42.3                & 0.82              \\
eng-npi             & 11.3             & 0.47          & 17.6             & 0.59           & 13.9                & 0.52             & 16.4                & 0.57              & 13.7                & 0.52             & 15.3                & 0.54              \\
eng-pol             & 23.5             & 0.70          & 23.5             & 0.70           & 24.9                & 0.70             & 24.2                & 0.70              & 24.4                & 0.70             & 23.3                & 0.70              \\
eng-rus             & 32.4             & 0.71          & 32.3             & 0.70           & 33.1                & 0.71             & 33.2                & 0.70              & 33.5                & 0.71             & 32.5                & 0.70              \\
eng-spa             & 29.8             & 0.74          & 28.9             & 0.74           & 29.2                & 0.74             & 29.6                & 0.74              & 30.2                & 0.74             & 29.5                & 0.74              \\
eng-swh             & 34.4             & 0.78          & 34.5             & 0.78           & 35.3                & 0.78             & 35.2                & 0.78              & 35.1                & 0.78             & 35.0                & 0.79              \\
eng-ukr             & 31.4             & 0.72          & 31.7             & 0.72           & 33.2                & 0.72             & 32.1                & 0.72              & 33.5                & 0.72             & 32.1                & 0.72              \\
eng-xho             & 21.9             & 0.73          & 21.7             & 0.73           & 22.3                & 0.73             & 24.2                & 0.75              & 22.9                & 0.73             & 23.4                & 0.74              \\
eng-zul             & 30.4             & 0.76          & 29.1             & 0.75           & 31.3                & 0.76             & 33.8                & 0.77              & 31.3                & 0.76             & 33.1                & 0.77              \\
est-eng             & 36.8             & 0.78          & 36.7             & 0.78           & 37.6                & 0.79             & 36.3                & 0.78              & 37.4                & 0.79             & 36.9                & 0.79              \\
fin-eng             & 34.5             & 0.78          & 35.1             & 0.78           & 35.6                & 0.78             & 35.1                & 0.78              & 35.8                & 0.78             & 35.0                & 0.78              \\
fra-eng             & 47.0             & 0.83          & 47.2             & 0.83           & 48.2                & 0.83             & 47.9                & 0.83              & 48.2                & 0.83             & 47.5                & 0.83              \\
guj-eng             & 29.9             & 0.76          & 29.7             & 0.76           & 31.5                & 0.76             & 30.7                & 0.76              & 31.4                & 0.76             & 29.0                & 0.75              \\
hin-eng             & 34.9             & 0.79          & 35.0             & 0.79           & 36.8                & 0.80             & 35.7                & 0.79              & 36.6                & 0.79             & 35.4                & 0.79              \\
ind-eng             & 42.2             & 0.81          & 41.4             & 0.81           & 43.8                & 0.81             & 42.8                & 0.81              & 43.4                & 0.81             & 43.0                & 0.81              \\
kaz-eng             & 1.5              & 0.46          & 2.8              & 0.48           & 1.3                 & 0.47             & 4.8                 & 0.50              & 3.1                 & 0.47             & 3.4                 & 0.47              \\
kir-eng             & 19.9             & 0.69          & 19.4             & 0.69           & 20.9                & 0.70             & 20.3                & 0.69              & 20.7                & 0.70             & 19.9                & 0.69              \\
lav-eng             & 35.4             & 0.79          & 35.4             & 0.78           & 36.6                & 0.79             & 36.2                & 0.78              & 36.7                & 0.79             & 35.4                & 0.79              \\
lit-eng             & 33.3             & 0.77          & 33.0             & 0.77           & 34.4                & 0.77             & 33.2                & 0.76              & 34.0                & 0.77             & 33.0                & 0.77              \\
mar-eng             & 28.2             & 0.75          & 27.9             & 0.75           & 30.3                & 0.76             & 28.6                & 0.75              & 30.2                & 0.76             & 28.2                & 0.75              \\
mkd-eng             & 41.9             & 0.81          & 42.6             & 0.81           & 43.8                & 0.82             & 42.8                & 0.81              & 43.2                & 0.81             & 42.9                & 0.81              \\
msa-eng             & 43.4             & 0.81          & 42.3             & 0.81           & 44.2                & 0.82             & 44.6                & 0.82              & 43.9                & 0.81             & 43.8                & 0.82              \\
npi-eng             & 29.2             & 0.75          & 29.6             & 0.75           & 31.7                & 0.76             & 29.7                & 0.75              & 31.7                & 0.76             & 29.6                & 0.75              \\
pol-eng             & 31.5             & 0.77          & 31.6             & 0.76           & 32.4                & 0.77             & 31.9                & 0.76              & 32.1                & 0.77             & 31.6                & 0.76              \\
rus-eng             & 36.9             & 0.79          & 36.6             & 0.79           & 38.1                & 0.79             & 37.6                & 0.79              & 38.0                & 0.79             & 36.9                & 0.79              \\
spa-eng             & 33.1             & 0.79          & 33.4             & 0.79           & 33.7                & 0.79             & 33.7                & 0.79              & 34.1                & 0.79             & 33.5                & 0.79              \\
swh-eng             & 40.2             & 0.79          & 40.7             & 0.79           & 42.0                & 0.79             & 40.8                & 0.79              & 41.9                & 0.79             & 40.6                & 0.79              \\
ukr-eng             & 38.6             & 0.80          & 39.0             & 0.80           & 40.5                & 0.80             & 39.6                & 0.80              & 40.4                & 0.80             & 39.4                & 0.80              \\
xho-eng             & 31.4             & 0.73          & 31.2             & 0.73           & 33.1                & 0.74             & 32.4                & 0.73              & 32.8                & 0.74             & 31.3                & 0.73              \\
zul-eng             & 34.8             & 0.76          & 34.4             & 0.76           & 36.0                & 0.76             & 37.1                & 0.77              & 36.2                & 0.76             & 36.2                & 0.76              \\ \bottomrule
\end{tabular}
    }
    \caption{Detailed performance of each translation direction for all models shown in Table~\ref{tab:main_res}}
    \label{tab:full_res}
\end{table*}
Table~\ref{tab:full_res} contains the detailed performance of all translation directions for all models reported in Table~\ref{tab:main_res}.

\section{Norm Analysis}\label{sec:norm_analysis}

The analysis in section~\ref{subsec:catastrophic} is limited to only measuring the lost information in the embeddings, to understand how learning rate scaling affects the weights other than embeddings, one natural extension is substituting other weights back to M20 as well. However, it leads to much worse performance for both variants as the inter-dependency among layers is impaired in this case. Therefore, we instead measure the amount that the weights of each encoder or decoder layer (denoted with $L_0, L_1,\ldots, L_{11})$ change in the latent space via the Frobenius norm. Following the notation in Figure~\ref{fig:lr_scale}, we measure how much the weight matrices $M_1$ and $M_2$ have changed from the original weight matrix $M$ ($\|M_1-M\|$ and $\|M_2-M\|$), as well as how much they differ from each other ($\|M_1-M_2\|$). We refer to the Frobenius norm in an encoder or a decoder layer with $\|\cdot\|_e$ and $\|\cdot\|_d$ respectively. The trend in Figure~\ref{fig:fro_norm} shows that applying learning rate scaling prevents $M_1$ from deviating too much from the original weights $M$ and at the same time pushes the new parameters to space farther from its initialization. This is in contrast with the smaller differences when not applying learning rate scaling. The left side of Figure~\ref{fig:fro_norm} indicates that even after the continual learning, $M_1$ and $M_2$ stay close to each other across the encoder layers, it is only when reaching the last several decoder layers do the two matrices demonstrate larger differences. Since we initialize $M_1$ and $M_2$ both based on $M$, having larger $\|M_1-M_2\|$ reduces the chance of both parts learning redundant information i.e., effectively using the additional parameters.

\section{Scaling learning rate by Fisher information}

Besides multiplying the learning rate for all old parameters with the same scaling factor, we also tried scaling the learning rate based on their Fisher Information. This is directly inspired by Elastic Weight Consolidation~\cite{EWC}, in which extra penalty is incurred when parameters crucial to the old tasks deviate too much from their original values. We plot the distribution of the per-token Fisher information of each parameter in Figure~\ref{fig:bleu_drop_mt25} (right). We further experiment with LR scaling over selected parameters that are supposed to be important for old tasks (Fisher information exceeds certain threshold based on Figure~\ref{fig:bleu_drop_mt25} (right)). Results in Table~\ref{tab:m12t25} (right) shows that among our attempted settings, scaling only part of the parameters based on Fisher information does not improve the overall performance. We conjecture that the performance could be improved if the Fisher Information is calculated on a larger set and that if we apply a piece-wise threshold function for scaling the learning rate of different parameters.

\section{Effect of LR scaling in alleviating catastrophic forgetting}
We include the spBLEU drop on the old languages after substituting the embeddings of Mt25 back to M20 in Figure~\ref{fig:bleu_drop_mt25}.

\end{document}